\documentclass[11pt]{article}

\usepackage[final]{acl}
\usepackage{multirow}
\usepackage{times}
\usepackage{latexsym}
\usepackage{amsmath}
\usepackage{booktabs}
\usepackage{tabularx}
\usepackage[T1]{fontenc}

\usepackage[utf8]{inputenc}
\usepackage{amssymb}
\usepackage{microtype}

\usepackage{inconsolata}

\usepackage{graphicx}

\title{Mechanistic Failure Modes of Weight Pruning in Transformer Language Models}

\author{Nabeelah Faumi  \\
  Informatics Institute of Technology\\
57, Ramakrishna Road, Colombo 06,\\
Sri Lanka\\
  \texttt{nabeelahfaumi@gmail.com } \\\And
  Deshan Sumanathilaka \\
  Informatics Institute of Technology\\
  57, Ramakrishna Road, Colombo 06,\\
 Sri Lanka\\
  \texttt{deshan.s@iit.ac.lk} \\}

\begin{document}
\maketitle
\begin{abstract}
Pruned language models are judged almost entirely by accuracy and perplexity, which assumes that preserved output behaviour implies preserved internal computation. We test this assumption with a factorial pruning study on GPT-2, crossing two importance criteria (magnitude, random), two component types (attention heads, MLP neurons) and three sparsity levels (30\%, 50\%, 70\%). Using the fully characterised indirect object identification (IOI) circuit as ground truth, we measure circuit integrity with activation patching, path patching and direct logit attribution. We find two qualitatively distinct failure modes. Magnitude head pruning ($\geq$50\%) \emph{zeroes} the circuit: output-facing heads are removed, and the model falls to near-chance. Magnitude neuron pruning at 30\% \emph{inverts} it: all 26 circuit heads remain intact, but Name Mover contributions flip sign, and the model systematically prefers the wrong name (17.7\% accuracy). Perplexity ranks it as less damaged than random head pruning (74.6 vs 140.1). A replication of the Greater-Than circuit shows that the critical component type depends on the circuit's architecture. These results show that output metrics and structural checks can both miss mechanistic damage, motivating circuit-level causal analysis as a complementary evaluation of compressed language models.

\end{abstract}

\section{Introduction}
Compression is now a default step in the language model deployment pipeline: pruning, quantisation, and distillation are routinely applied to reduce memory footprint and inference cost, often as a black-box post-processing stage after training \citep{ding2023,bai2024}. Because these methods are judged almost entirely by their effect on downstream benchmarks, accuracy and perplexity, they are treated as behaviourally neutral whenever those numbers hold up. This paper asks whether that judgment is safe to make.
The concern is not merely academic. A model can produce a correct answer for the wrong reason, and the pathways that mechanistic interpretability has isolated in unpruned models, such as the IOI circuit \citep{wang2022}, give a concrete setting in which "correct reason" can actually be checked, rather than assumed. If pruning is capable of severing or inverting the specific components responsible for a behaviour while leaving redundant computation to paper over the loss at the output layer, then accuracy-based compression audits are measuring the wrong thing, and models cleared for deployment on that basis may carry mechanistic damage that only surfaces under distribution shift, adversarial input, or further fine-tuning.
This risk is what our factorial pruning study is designed to expose. Rather than treating pruning as a single method to be validated once, we vary both \emph{what} is removed (attention heads vs. MLP neurons) and \emph{how} it is selected (magnitude-based vs. random), across a range of sparsities, and ask whether these choices leave distinguishable fingerprints on a well-characterised circuit's internal structure, independent of whatever the accuracy metric reports.

Our contributions are twofold. First, we identify two qualitatively distinct mechanistic failure modes of structured pruning in Transformer language models: circuit zeroing and circuit inversion. We show that MLP-neuron pruning can invert an otherwise structurally intact circuit, transforming a functioning circuit into an actively misleading circuit that produces systematic rather than random errors. Second, we demonstrate that standard behavioural and structural pruning evaluations can miss such mechanistic failures: for example, perplexity may rank an inverted model as less damaged, while checking whether circuit components remain present can falsely suggest that the circuit is intact. Our results therefore motivate circuit-level causal analysis as a complementary evaluation of compressed language models.

In the following sections, we review related work, outline our experimental setup, present our results, and conclude with a discussion and conclusion.

\section{Background and Related Work}

\subsection{Mechanistic Interpretability and Circuit Analysis}
Mechanistic interpretability reverse-engineers neural networks by identifying circuits, interpretable components that implement human-understandable algorithms \citep{bereska2024,cammarata2020}. In transformers, circuit analysis decomposes the residual stream into attention-head and MLP contributions, then uses causal interventions to test which are necessary and sufficient for a behaviour \citep{wang2022,meng2022,geiger2021}. The Indirect Object Identification (IOI) task requires completing sentences such as \emph{``When Mary and John went to the store, John gave a drink to \_\_ ,''} with the indirect object (IO, \emph{Mary}) rather than the subject (S, \emph{John}). \citet{wang2022} identified a 26-head circuit in GPT-2: Previous Token (PT) heads copy token positions; Duplicate Token (DT) and Induction (IND) heads detect repeated names; S-Inhibition (S-INH) heads suppress the subject; Name Mover (NM) heads copy the IO to the output; Negative Name Mover (NNM) heads suppress the IO signal as a self-correcting brake; and Backup Name Mover (BNM) heads compensate when NM heads fail. This gives a well-characterised ground truth against which compression-induced changes can be measured.

We use two complementary tools: \emph{activation patching}, which replaces a head's activations with those from a corrupted run to measure causal necessity, and \emph{Direct Logit Attribution} (DLA), a non-interventional decomposition of the logit difference into per-head contributions that measures direct contribution \citep{goldowsky2023,wang2022}.

\subsection{Weight Pruning}

Pruning assumes that many parameters are redundant: unstructured pruning removes individual weights, while structured pruning removes entire neurons, filters, or attention heads \citep{ding2023,bai2024}. Magnitude-based pruning, the most widely used criterion, removes low-magnitude weights \citep{han2015}; random pruning serves as a controlled baseline. From an interpretability perspective, the key distinction is between attention heads, which implement relational computation across token positions, and MLP neurons, which store features, knowledge and position-sensitive representations that downstream heads depend on \citep{geva2021,rai2024}. This motivates our $2\times 2$ factorial design.

\subsection{Pruning and Circuit Integrity: Prior Work}
Attribution-guided pruning (Layer-wise Relevance Propagation, Integrated Gradients, gradient-based attribution) preferentially retains task-relevant subgraphs \citep{hatefi2025,chhabra2025}, and Attribution-based Parameter Decomposition shows that interpretable circuit components can survive substantial size reduction \citep{braun2025}. In safety-aligned models, pruning may disproportionately erode safety-relevant circuits \citep{chhabra2025}. However, no prior work has characterised whether different pruning strategies produce qualitatively distinct circuit failures or shown that these failures are invisible to accuracy evaluation. We address both gaps.

\section{Experimental Setup}
 
\subsection{Base Model}
All experiments use GPT-2 \citep{radford2019}: 117M parameters, 12 transformer layers, 12 attention heads per layer, and a hidden dimension of 768. GPT-2 is selected because it is the canonical platform for mechanistic interpretability research, with a fully characterised IOI circuit providing a ground-truth benchmark. All compressed variants are derived from a single pre-trained checkpoint with no weight updates after pruning, ensuring cross-variant comparisons are made at matched starting conditions. All experiments are implemented using TransformerLens \citep{nanda2022}, which provides hook-based access to residual stream activations, attention head outputs, and MLP intermediate states.
 
\subsection{Pruning Variants}
 
Twelve pruned variants are created by crossing two importance criteria (magnitude, random), two component types (attention heads, MLP neurons), and three sparsity levels (30\%, 50\%, 70\%), yielding a $2 \times 2 \times 3$ factorial design.
 
\paragraph{Attention head pruning.}
Heads are pruned by zeroing all weight slices corresponding to a given head in the \texttt{c\_attn} projection matrix (Q, K, and V columns) and the corresponding rows of the \texttt{c\_proj} output projection. For magnitude pruning, each head is scored by the L2-norm of its Q-weight slice, and the $k$ lowest-scoring heads across the entire model are removed. Random pruning assigns importance scores uniformly at random and removes a fixed proportion of heads per layer.
 
\paragraph{MLP neuron pruning.}
Neurons are pruned by zeroing the corresponding columns of the first weight matrix
($W_{\text{in}}$) and rows of the second weight matrix ($W_{\text{out}}$) within each MLP
sub-layer.
For magnitude pruning, neurons are scored by the L2 norm of their $W_{\text{in}}$ column
and the lowest-scoring neurons are removed globally.
Random pruning removes a uniformly sampled fraction of neurons from each layer independently.
 
\subsection{Evaluation Datasets and Tasks}
 
Sentences follow the ABB/ABC design of \citep{wang2022}, built from 15 templates , 14 objects and 28 single-token names, and each is paired with a corrupted counterpart in which the second subject is replaced by a foil name to support activation patching. The primary metric is the mean logit difference $ld = \ell_{\text{IO}} - \ell_{\text{S}}$: the difference in log-probability between the indirect object name and the subject name at the prediction position. Positive values indicate correct IO prediction; negative values indicate the model assigns a higher probability to the subject. Circuit analysis uses activation patching across all 26 IOI circuit heads and DLA decomposed by functional class. General language modelling performance is evaluated on TinyStories \citep{li2023} via perplexity over 1000 sequences.
 
\subsection{Implementation Details}
 
All experiments use Python 3.10 with TransformerLens \citep{nanda2022} and HuggingFace Transformers \citep{wolf2020}. LayerNorm folding is explicitly disabled (\texttt{fold\_ln=False}) to preserve activation scales. DLA is computed using a linear approximation of the final LayerNorm centred on the full residual stream, ensuring component contributions sum to the observed logit difference within numerical precision. Sparsity fractions are verified after TransformerLens conversion to ensure zero-valued entries are correctly propagated.
All experiments are run on an NVIDIA L4/T4 GPU (Google Colab) with fixed random seed 42.
 
\section{Results}
 
\subsection{Behavioural Baseline}
 
GPT-2 achieves a mean IOI logit difference of $ld = 2.942 \pm 1.797$ and IOI accuracy of 94.8\% on the 300-sentence evaluation set. Perplexity on TinyStories is 11.54, serving as the reference baseline throughout. Table~\ref{tab:performance} reports performance across all compressed variants; Table~\ref{tab:dla} summarises the failure modes; Tables~\ref{tab:patching} and \ref{tab:dla} give the head-level mechanistic evidence. Perplexity and representational-fidelity results for the same variants are given in Section~\ref{sec:fidelity}.
 
\begin{table*}[ht]
\centering
\small
\caption{IOI task performance across all compressed models.
Degradation $= (\mu_{\text{base}} - \mu_{\text{comp}}) / \mu_{\text{base}}$; values $>100\%$ indicate a negative mean logit difference.}
\label{tab:performance}
\begin{tabular}{llrrr}
\toprule
\textbf{Method} & \textbf{Model} & \textbf{Mean LD} & \textbf{Acc (\%)} & \textbf{Degrad. (\%)} \\
\midrule
Base & GPT-2 Small (FP32) & $2.942$ & $94.8$ & ,  \\
\midrule
\multirow{4}{*}{Pruning 30\%}
  & Mag.\ Neurons  & $-2.475$ & $17.7$ & $184.1$ \\
  & Rand.\ Neurons & $-0.896$ & $35.6$ & $130.5$ \\
  & Mag.\ Attn     & $0.200$  & $51.4$ & $93.2$  \\
  & Rand.\ Attn    & $-0.706$ & $36.4$ & $124.0$ \\
\midrule
\multirow{4}{*}{Pruning 50\%}
  & Mag.\ Neurons  & $-1.318$ & $33.7$ & $144.8$ \\
  & Mag.\ Attn     & $-0.199$ & $45.7$ & $106.8$ \\
  & Rand.\ Neurons & $-0.502$ & $38.2$ & $117.1$ \\
  & Rand.\ Attn    & $-0.593$ & $31.0$ & $120.2$ \\
\midrule
\multirow{4}{*}{Pruning 70\%}
  & Mag.\ Neurons  & $-0.692$ & $42.2$ & $123.5$ \\
  & Mag.\ Attn     & $-0.113$ & $47.9$ & $103.8$ \\
  & Rand.\ Neurons & $-0.038$ & $47.9$ & $101.3$ \\
  & Rand.\ Attn    & $-0.242$ & $41.6$ & $108.2$ \\
\bottomrule
\end{tabular}%
\end{table*}
\subsection{Failure Mode I: Zeroing}
 
Magnitude attention head pruning at 50\% and 70\% sparsity physically eliminates the output-facing heads of the IOI circuit. At 50\% sparsity, DLA scores for all four major circuit classes, NM, NNM, S-INH, and BNM, fall to exactly zero (Table~\ref{tab:dla}). Weight norm verification confirms that circuit heads have been physically removed: pruned circuit heads collapse to uniform attention patterns ($\text{IO\_attn} = 0.067$, $\text{S\_attn} = 0.133$). 
At 30\% sparsity, 10 of 26 circuit heads are removed; the surviving NM heads (L9H6 and L9H9) retain partial IO-preferring attention (L9H9: $\text{IO\_attn} = 0.536$, $\text{S\_attn} = 0.083$), explaining the partial 9.2\% NM retention at this sparsity.
Despite the elimination of the circuit, the model retains 45.7\%--47.9\% IOI accuracy at 50\%--70\% sparsity. This residual accuracy derives from non-circuit computation: early-circuit heads and MLP layers are not targeted by magnitude attention head pruning (which removes lower-norm output-facing heads first), and their intact language-modelling signal incidentally favours the IO token over the S token on most sentences.
Zeroing is the most interpretable and, from a safety perspective, the least dangerous failure mode: the circuit has been structurally removed, the cause of degraded performance is straightforwardly attributable to physical elimination of circuit components, and the model makes near-chance predictions rather than actively misdirecting.
 
\subsection{Failure Mode II: Inversion}
Magnitude neuron pruning at 30\% sparsity produces a qualitatively distinct and more dangerous failure. All 26 IOI circuit heads remain structurally intact (weight norms all $>25$), yet DLA analysis reveals that major circuit classes have flipped sign and actively harm correct performance.
At 30\% sparsity, the circuit exhibits severe inversion as Name Mover (NM) and Backup Name Mover (BNM) heads flip to negative DLA contributions, retaining $-30.8\%$ ($-1.102$ DLA) and an extreme $-119.9\%$ of their base function, respectively. Simultaneously, Negative Name Mover (NNM) heads lose their correct suppressive role ($-23.0\%$ retention), meaning the output mechanism actively harms predictive accuracy.
The result is a mean IOI logit difference of $-2.475$ and accuracy of 17.7\%: the model is actively and confidently predicting the wrong name.
The mechanism is \emph{context poisoning}. Magnitude pruning preferentially removes the largest MLP neurons, which implement the position-specific representations and name-identity encodings that attention heads depend on to locate the correct indirect object. The NM heads remain functional as copying mechanisms, but the corrupted residual stream directs them to the wrong entity. Attention pattern analysis confirms this: L9H6 shows $\text{IO\_attn} = 0.051$, $\text{S\_attn} = 0.650$; L9H9 shows $\text{IO\_attn} = 0.123$, $\text{S\_attn} = 0.600$. The heads are operational but pointed at the subject. 
At 30\% sparsity, the residual stream is corrupted enough to invert the NM signal while still providing sufficient discriminability for consistent (though wrong) predictions.
At 70\% sparsity, the residual stream is so broadly degraded that attention patterns lose discriminability, weakening both correct and inverted signals and moving the logit difference back towards zero. The 30\% model is \emph{strongly wrong}; the 70\% model is \emph{weakly wrong}, a pattern invisible to accuracy evaluation, which shows only modestly reduced accuracy at 30\% relative to the catastrophic underlying circuit failure.

\begin{table*}[ht]
\centering
\caption{DLA retention (\%) by circuit class across compressed models. Negative values indicate sign inversion (the class actively harms correct IO prediction). $\approx 0$ denotes exact zero from structurally eliminated heads.}
\label{tab:dla}
\small
\begin{tabular}{llrrrrl}
\toprule
\textbf{Method} & \textbf{Model} & \textbf{S-INH} & \textbf{NM} & \textbf{NNM}
  & \textbf{BNM} & \textbf{Failure Mode} \\
\midrule
\multirow{4}{*}{Pruning 30\%}
  & Mag.\ Attn  & $17.6$  & $9.2$    & $\approx 0$ & $-2.2$    & Zeroing (partial) \\
  & Rand.\ Attn & $50.2$  & $-8.9$   & $-11.5$     & $-59.8$   & Inversion (partial) \\
  & Mag.\ Neu   & $21.2$  & $-30.8$  & $-23.0$     & $-119.9$  & Inversion \\
  & Rand.\ Neu  & $26.8$  & $-8.5$   & $-3.7$      & $-44.0$   & Inversion (partial) \\
\midrule
\multirow{4}{*}{Pruning 50\%}
  & Mag.\ Attn  & $-7.1$  & $\approx 0$ & $\approx 0$ & $\approx 0$ & Zeroing \\
  & Rand.\ Attn & $10.0$  & $-5.3$   & $-5.4$      & $-37.3$   & Inversion (partial) \\
  & Mag.\ Neu   & $15.8$  & $-10.9$  & $-3.5$      & $-35.7$   & Inversion (partial) \\
  & Rand.\ Neu  & $13.0$  & $-4.0$   & $-0.9$      & $-8.7$    & Inversion \\
\midrule
\multirow{4}{*}{Pruning 70\%}
  & Mag.\ Attn  & $\approx 0$ & $\approx 0$ & $\approx 0$ & $\approx 0$ & Zeroing \\
  & Rand.\ Attn & $13.0$  & $-2.1$   & $-0.9$      & $-7.2$    & Inversion (mild) \\
  & Mag.\ Neu   & $11.3$  & $-4.4$   & $-0.8$      & $-2.5$    & Inversion (mild) \\
  & Rand.\ Neu  & $6.9$   & $-1.2$   & $-0.0$      & $-1.3$    & Strong Dampening\\
\bottomrule
\end{tabular}%
\end{table*}

\subsection{High-Sparsity Limit of Inversion}
Random neuron pruning at 70\% sparsity produces the weakest circuit signature in the study. No class shows a strong sign flip: NM ($-1.2\%$), NNM ($-0.0\%$) and BNM ($-1.3\%$) are indistinguishable from zero, and S-Inhibition retains only 6.9\% of its baseline contribution (Table~\ref{tab:dla}). Accuracy (47.9\%) and mean logit difference ($-0.038$) match the zeroing profile at equivalent sparsity (Table~\ref{tab:performance}), but the cause differs: neuron pruning does not touch the attention heads, so the circuit is not absent; its directional signal has faded. Activation patching shows a partial exception: S-Inhibition retains a moderate causal effect ($d=0.270$), whereas NNM retains its paradoxical negative effect ($d=-0.329$). The same trajectory appears across sparsity levels, with the logit difference decreasing from $-0.896$ at 30\% to $-0.502$ and $-0.038$ at 50\% and 70\%, respectively. We therefore treat this regime as the terminal, noise-dominated end of inversion, in which context corruption is so broad that both the correct and the inverted signals fade, rather than as an independent mechanism.

\subsection{Perplexity and Representational Fidelity}
\label{sec:fidelity}

Table~\ref{tab:fidelity} places the circuit-level failures in the context of conventional and representational measures: TinyStories perplexity, hidden-state cosine similarity, and linear CKA between each pruned model and the base model, reported overall and for early (L0--L3) and late (L8--L11) layers.

\begin{table*}[ht]
\centering
\caption{TinyStories perplexity and layer-wise representational similarity to the base model (base perplexity 11.54). Early $=$ L0--L3, Late $=$ L8--L11.}
\label{tab:fidelity}
\small
\begin{tabular}{llrrrrr}
\toprule
\textbf{Spar.} & \textbf{Model} & \textbf{PPL} & \textbf{Cos.} & \textbf{CKA} & \textbf{CKA (early)} & \textbf{CKA (late)} \\
\midrule
\multirow{4}{*}{30\%}
  & Mag.\ Neu   & $74.61$  & $.856$ & $.698$ & $.946$ & $.508$ \\
  & Rand.\ Neu  & $102.09$ & $.917$ & $.825$ & $.993$ & $.553$ \\
  & Mag.\ Attn  & $23.37$  & $.980$ & $.941$ & $.998$ & $.848$ \\
  & Rand.\ Attn & $140.10$ & $.880$ & $.801$ & $.986$ & $.648$ \\
\midrule
\multirow{4}{*}{50\%}
  & Mag.\ Neu   & $262.28$ & $.813$ & $.577$ & $.900$ & $.366$ \\
  & Rand.\ Neu  & $300.49$ & $.897$ & $.791$ & $.977$ & $.520$ \\
  & Mag.\ Attn  & $41.24$  & $.977$ & $.917$ & $.998$ & $.773$ \\
  & Rand.\ Attn & $140.65$ & $.894$ & $.817$ & $.987$ & $.647$ \\
\midrule
\multirow{4}{*}{70\%}
  & Mag.\ Neu   & $18{,}934$ & $.703$ & $.500$ & $.906$ & $.267$ \\
  & Rand.\ Neu  & $1{,}126$  & $.850$ & $.646$ & $.970$ & $.260$ \\
  & Mag.\ Attn  & $83.79$    & $.957$ & $.851$ & $.979$ & $.608$ \\
  & Rand.\ Attn & $538.39$   & $.781$ & $.705$ & $.981$ & $.333$ \\
\bottomrule
\end{tabular}%
\end{table*}

\paragraph{Perplexity does not track circuit damage.}
Magnitude attention head pruning has the lowest perplexity at every sparsity (23.37 at 30\%, against 140.10 for random attention heads at the same sparsity), and neuron pruning is far more damaging than head pruning at matched sparsity, with magnitude neurons at 70\% reaching a perplexity of 18{,}934. At 30\% sparsity, however, perplexity ranks magnitude neuron pruning (74.61) ahead of random attention head pruning (140.10), even though the former produces the worst IOI logit difference in the study ($-2.475$ against $-0.706$; Table~\ref{tab:performance}). The most strongly inverted circuit, therefore, belongs to a model that a perplexity audit would rank as the less damaged of the two.

\paragraph{Component type and depth.}
Representational fidelity separates component types more sharply than selection criteria. At 70\% sparsity, magnitude attention head pruning retains a CKA of 0.851, the highest of any pruned model at that sparsity, whereas magnitude neuron pruning retains only 0.500. For neuron pruning, cosine similarity (0.703) and CKA (0.500) diverge at 70\%, indicating that the principal axes of variation in the MLP output space collapse while mean activation directions are partially preserved. All strategies preserve early layers (CKA $\geq 0.900$ in every condition) while late-layer CKA never exceeds 0.848 and falls as low as 0.260, so the representational bottleneck under pruning is a late-layer phenomenon, which is where the IOI output-stage heads sit.

\paragraph{Criterion reversal for neurons.}
For attention heads, magnitude selection preserves more structure than random selection (CKA 0.917 against 0.817 at 50\%). For neurons, the ordering reverses at low sparsity: random pruning at 30\% reaches a CKA of 0.825 against 0.698 for magnitude pruning. This is consistent with the context-poisoning account in Inversion: magnitude selection removes the highest-weight neurons, creating localised representational voids, whereas random selection spreads the damage more evenly.

\subsection{Attention Diffusion and Sharpening}
\label{sec:attention}
 
We track head-averaged attention statistics (entropy, effective support, first-token mass, and head cosine similarity to the base model; full results in Table~\ref{tab:attention}.
 
\paragraph{Head pruning diffuses attention.}
Pruning entire heads produces systematic attention diffusion under both criteria. At 70\% sparsity, entropy rises monotonically from 2.013 nats to 3.561 (magnitude) and 3.536 (random), effective support grows from 11.4 to roughly 51 tokens (a 4.5-fold increase), and first-token attention collapses from 0.421 to 0.098 (magnitude) and 0.146 (random), destroying the attention sink. This is a structural consequence of zeroing a head's weights, which yields near-uniform softmax outputs regardless of how the head was selected, and it matches the uniform IO and S attention measured on the eliminated circuit heads ($\text{IO\_attn}=0.067$, $\text{S\_attn}=0.133$). Head cosine similarity falls heterogeneously ($\bar s=0.524$ and $0.482$ at 70\%), reflecting damage concentrated in the removed heads.
 
\paragraph{Neuron pruning does not diffuse attention.}
Neuron pruning leaves attention heads untouched, and its signature is opposite or weak. Random neuron pruning \emph{sharpens} attention: entropy is below base at 30\% and 70\% (1.638 and 1.478 nats), effective support falls to 9.25 and 7.80 tokens, and first-token attention rises to 0.521 and 0.552. The 50\% condition (2.050 nats) interrupts this trend and is likely a stochastic or threshold effect. Magnitude neuron pruning produces only a mild entropy increase ($\Delta H=+0.173$ to $+0.394$). A plausible mechanism is a compensatory collapse of the residual stream: with fewer MLP neurons, hidden states become less diverse and attention concentrates on contextually dominant positions. At matched sparsity, random neuron pruning also preserves head routing better than random head pruning ($\bar s=0.894$ vs.\ $0.704$ at 30\%; $0.819$ vs.\ $0.482$ at 70\%).
 
\paragraph{Implication for detection.}
Zeroing is visible in aggregate attention statistics, whereas the inverted model at 30\% magnitude neuron pruning ($\Delta H=+0.173$) looks comparatively intact, even though its Name Mover heads attend sharply to the wrong token ($\text{S\_attn}=0.650$ for L9H6). Aggregate entropy cannot distinguish a head that attends correctly from one that attends confidently to the subject, which is why head-level causal analysis is needed.

\subsection{Greater-Than Circuit Replication}
\label{sec:gt}

To test whether the taxonomy is specific to IOI, we apply the same twelve pruned variants to the Greater-Than (GT) circuit of \citet{hanna2023}. 
Each prompt has the form ``The \emph{noun} lasted from the year XXYY to the year XX'' and requires a completion YY$'>$YY. The circuit comprises six heads in four classes: year encoding (L0H1, L1H5), year difference (L9H1), comparison (L10H7), and output computation (L11H0, L11H10). Unlike IOI, the GT computation is MLP-dominant: the largest direct logit attributions in GPT-2 come from the layer 9, 10 and 8 MLPs (0.867, 0.863 and 0.557), with L9H1 (0.551) the first attention component. The metric is the GT logit difference $\Delta GT$, the mean logit of valid completions minus the mean logit of invalid ones. 
The base model achieves $\Delta GT = 2.855 \pm 1.942$ and an argmax accuracy of 94.0\%.

\begin{table*}[ht]
\centering
\caption{GT logit difference, accuracy and degradation under pruning. Degradation $>100\%$ indicates a negative mean logit difference. Base $\Delta GT=2.855$.}
\label{tab:gt}
\small
\begin{tabular}{llrrr}
\toprule
\textbf{Spar.} & \textbf{Model} & \textbf{Mean $\Delta GT$} & \textbf{Acc (\%)} & \textbf{Degrad. (\%)} \\
\midrule
\multirow{4}{*}{30\%}
  & Mag.\ Neu   & $-0.580 \pm 0.643$ & $0.2$  & $120.3$ \\
  & Rand.\ Neu  & $0.421 \pm 1.286$  & $33.8$ & $85.3$  \\
  & Mag.\ Attn  & $1.197 \pm 1.046$  & $66.1$ & $58.1$  \\
  & Rand.\ Attn & $1.066 \pm 1.701$  & $49.9$ & $62.6$  \\
\midrule
\multirow{4}{*}{50\%}
  & Mag.\ Neu   & $-0.369 \pm 0.730$ & $0.0$  & $112.9$ \\
  & Rand.\ Neu  & $-0.416 \pm 0.349$ & $0.0$  & $114.6$ \\
  & Mag.\ Attn  & $0.921 \pm 0.959$  & $53.5$ & $67.7$  \\
  & Rand.\ Attn & $0.214 \pm 0.818$  & $1.0$  & $92.5$  \\
\midrule
\multirow{4}{*}{70\%}
  & Mag.\ Neu   & $-0.424 \pm 0.881$ & $0.0$  & $114.8$ \\
  & Rand.\ Neu  & $-0.118 \pm 0.416$ & $0.0$  & $104.1$ \\
  & Mag.\ Attn  & $-0.091 \pm 0.660$ & $17.7$ & $103.2$ \\
  & Rand.\ Attn & $0.324 \pm 0.469$  & $8.2$  & $88.7$  \\
\bottomrule
\end{tabular}%
\end{table*}

\paragraph{Neuron pruning is the most destructive.}
Magnitude neuron pruning inverts $\Delta GT$ at every sparsity level, with accuracy at or below 0.2\%. Random neuron pruning retains a small positive logit difference at 30\% (0.421, 33.8\% accuracy) but inverts at 50\% and 70\%, and accuracy reaches exactly 0.0\% by 50\% under both criteria. Because the GT circuit relies on distributed MLP computation, the surviving attention heads cannot compensate for it. Magnitude selection is again more damaging than random selection for neurons at 30\% (degradation 120.3\% against 85.3\%), plausibly because it removes the highest-activating neurons most likely to carry the ordinal comparison signal.

\paragraph{Attention head pruning degrades gradually.}
Magnitude attention head pruning retains meaningful performance (accuracies of 66.1\%, 53.5\%, and 17.7\%) and crosses the zero logit difference only at 70\%. Random attention head pruning degrades more sharply, with accuracy falling to 1.0\% at 50\%, and the gap between the two criteria widens with sparsity (degradation 58.1\% against 62.6\% at 30\%, and 67.7\% against 92.5\% at 50\%).

\paragraph{Circuit-level evidence.}
Head-level activation patching shows that L10H7 (comparison) and L11H0 and L11H10 (output) are the heads most affected by attention pruning. L9H1 (year difference) shows \emph{increased} patching scores in pruned models, consistent with compensation as output heads fail and L9H1 absorbs a larger causal load, whereas in neuron-pruned models, the same heads show negative effect sizes because no recoverable signal is computed. Path patching (Table~\ref{tab:gtpaths}) shows CMP$\to$OUT and YE$\to$YD as the most consistently damaged paths. At 70\% random attention pruning, all four paths degrade by more than 100\%, indicating complete circuit dissolution, whereas under magnitude attention pruning at 70\%, YD$\to$CMP ($-90.2\%$) and YD$\to$OUT ($-32.4\%$) show negative degradation, reflecting compensatory re-routing as the output heads fail.

\paragraph{Relation to the IOI taxonomy.}
GT reproduces the asymmetry between component types, but with the roles reversed. For IOI the circuit is carried by attention heads, and neuron pruning is dangerous because it corrupts the context those heads read; for GT the circuit is carried by MLPs, so neuron pruning removes the computation itself. Which component type is critical, therefore, depends on the circuit's architecture, but in both circuits, the choice of component type, rather than sparsity, determines the failure.

\begin{table*}[ht]
\centering
\caption{GT residual path patching: mean degradation (\%) and Cohen's $d$. Negative degradation indicates increased causal potency in the compressed model.}
\label{tab:gtpaths}
\small
\begin{tabular}{llrrrrrrrr}
\toprule
 & & \multicolumn{2}{c}{YE$\to$YD} & \multicolumn{2}{c}{YD$\to$CMP} & \multicolumn{2}{c}{YD$\to$OUT} & \multicolumn{2}{c}{CMP$\to$OUT} \\
\cmidrule(lr){3-4}\cmidrule(lr){5-6}\cmidrule(lr){7-8}\cmidrule(lr){9-10}
\textbf{Spar.} & \textbf{Model} & Deg. & $d$ & Deg. & $d$ & Deg. & $d$ & Deg. & $d$ \\
\midrule
\multirow{4}{*}{30\%}
  & Mag.\ Neu   & $49.9$  & $-.55$  & $-11.4$ & $.06$  & $22.6$  & $-.17$  & $54.4$  & $-.67$  \\
  & Rand.\ Neu  & $44.4$  & $-.56$  & $-18.8$ & $.12$  & $17.4$  & $-.15$  & $49.0$  & $-.68$  \\
  & Mag.\ Attn  & $38.4$  & $-1.16$ & $1.1$   & $-.02$ & $27.0$  & $-.57$  & $42.9$  & $-1.49$ \\
  & Rand.\ Attn & $22.6$  & $-.26$  & $39.0$  & $-.22$ & $41.2$  & $-.33$  & $15.4$  & $-.20$  \\
\midrule
\multirow{4}{*}{50\%}
  & Mag.\ Neu   & $62.2$  & $-.71$  & $19.4$  & $-.11$ & $43.9$  & $-.34$  & $65.3$  & $-.82$  \\
  & Rand.\ Neu  & $44.4$  & $-.56$  & $-18.8$ & $.12$  & $17.4$  & $-.15$  & $49.0$  & $-.68$  \\
  & Mag.\ Attn  & $39.4$  & $-.94$  & $-31.3$ & $.38$  & $8.6$   & $-.13$  & $45.1$  & $-1.25$ \\
  & Rand.\ Attn & $76.3$  & $-1.40$ & $70.4$  & $-.74$ & $79.0$  & $-1.05$ & $78.1$  & $-1.54$ \\
\midrule
\multirow{4}{*}{70\%}
  & Mag.\ Neu   & $54.9$  & $-.55$  & $4.7$   & $-.02$ & $33.6$  & $-.22$  & $59.2$  & $-.66$  \\
  & Rand.\ Neu  & $88.7$  & $-1.03$ & $76.6$  & $-.45$ & $83.5$  & $-.65$  & $89.5$  & $-1.14$ \\
  & Mag.\ Attn  & $12.6$  & $-.22$  & $-90.2$ & $.92$  & $-32.4$ & $.41$   & $20.5$  & $-.41$  \\
  & Rand.\ Attn & $111.2$ & $-1.42$ & $124.4$ & $-.84$ & $117.0$ & $-1.02$ & $109.9$ & $-1.53$ \\
\bottomrule
\end{tabular}%
\end{table*}

\section{Discussion}
 
\subsection{Component Type and Selection Criterion Jointly Determine Failure Mode}

Component type matters more than sparsity. Magnitude attention pruning at 70\% (zeroing) retains 47.9\% IOI accuracy, whereas magnitude neuron pruning at 30\% (inversion) retains only 17.7\%, despite zeroing far fewer parameters.
Within a component type, however, the selection criterion changes the failure signature (Table~\ref{tab:dla}). At 30\% sparsity, magnitude attention pruning gives the expected zeroing profile (S-INH 17.6\%, NM 9.2\%, NNM $\approx 0$, BNM $-2.2\%$). Random attention pruning at the same sparsity instead flips NM ($-8.9\%$), NNM ($-11.5\%$) and especially BNM ($-59.8\%$), a partial-inversion signature otherwise associated with neuron pruning.
Path patching suggests why (Table~\ref{tab:path-full}). Low-norm heads are disproportionately the circuit's own late-layer output heads, so magnitude selection is close to surgical: S-INH$\to$NM degrades most (87.4\% vs.\ 36.6--66.5\% upstream). Random selection also removes upstream PT, DT and IND heads that feed the surviving Name Movers, and degrades all four pathways (PT$\to$IND 201.3\%, DT$\to$S-INH 122.0\%, IND$\to$NM 143.2\%, S-INH$\to$NM 91.2\%).
Inversion thus arises whenever the context Name Movers read is damaged while the Name Movers survive, whether via neurons or via randomly removed upstream heads.
A practitioner choosing between magnitude and random head pruning is therefore choosing between a self-announcing failure and a silent one, even though the pruned component type is identical.
One caveat: for neuron pruning, patching effect sizes fall with sparsity (magnitude: $d=0.135$ at 30\% to $0.059$ at 70\%) while performance worsens, indicating global disruption in which patching loses sensitivity, so $d$ should not be read as damage magnitude across sparsity levels.
 
\subsection{Why Inversion Is the Most Dangerous Failure Mode}
A zeroed model fails transparently: at 70\% magnitude attention pruning every DLA class is exactly zero and attention over IO and S is near-uniform (0.067 vs.\ 0.133), so the residual accuracy comes from non-circuit computation.
Inversion leaves no such structural trace. Under 30\% magnitude neuron pruning, all 26 circuit heads remain intact (weight norms $>25$), so an audit checking whether the circuit still exists would pass it. Yet the output mechanism has been redirected: NM contribution falls from $+3.576$ to $-1.102$, BNM from $+0.427$ to $-0.511$ ($-119.9\%$, the most extreme inversion in the study), and S-INH retains 21.2\%. The mean logit difference is $-2.475$ with 17.7\% accuracy, the worst of all twelve variants, which is a systematic preference for the subject rather than undirected guessing.
Perplexity misranks this model: at 30\% sparsity it scores 74.61 against 140.10 for random attention pruning, yet its logit difference is far worse ($-2.475$ vs.\ $-0.706$).
Degradation is also non-monotonic in sparsity (184.1\% at 30\%, 123.5\% at 70\%), because heavier pruning degrades the residual stream until attention no longer discriminates between candidates. The model is thus most misdirected at the sparsity practitioners are most likely to consider conservative.
Structural checks miss behavioural inversion, while accuracy and perplexity cannot distinguish uncertainty from systematic error. For safety mechanisms, the pruning method could determine whether failure merely reduces capability or reverses the intended behaviour.
 
\subsection{Structural Weak Points}
Two components are disrupted across conditions. The S-INH$\to$NM pathway is degraded in every condition we report (path-patching $d=0.102$--$0.365$; Table~\ref{tab:path-full}), even under random neuron pruning at 30\%, where all other pathways show reorganisation rather than damage. Because S-Inhibition is distributed across the MLP and attention components at layers 7--8, it is exposed to any pruning strategy, regardless of the target.
NNM heads show negative patching $d$ in all twelve conditions ($-0.205$ to $-0.539$; Table~\ref{tab:patching}), i.e.\ their suppressive role is decoupled or inverted rather than merely weakened. We hypothesise that this reflects the precision NNM heads need to target the subject token specifically, compared with the more distributed IO-copying signal of NM heads; we do not test this directly.

\section{Conclusion}
We demonstrate that weight pruning disrupts transformer circuits through two distinct failure modes, driven primarily by component type rather than sparsity level. Magnitude attention head pruning eliminates output-facing heads outright, producing undirected, near-chance failure. Magnitude neuron pruning instead corrupts the MLP context that attention heads depend on, producing confident misdirection that accuracy metrics obscure. At high sparsity, random neuron pruning shows both correct- and incorrect-sign contributions fading toward zero together, a terminal, noise-dominated regime rather than a stable, independently characterisable third mechanism.  

\section*{Limitations}
 
\paragraph{Model and scale.}
All experiments use GPT-2 small (117M parameters). Larger models have more numerous heads and more distributed, redundant circuits, so the failure modes we report may not transfer directly, and replication at larger scales (e.g.\ GPT-2 Medium/Large or Pythia) is the most immediate extension.
 
\paragraph{Circuit coverage and data.}
We analyse two circuits, IOI and Greater-Than, on synthetically generated prompts (300 IOI sentences). Synthetic data is standard in circuit analysis but limits ecological validity, and we do not test whether circuit damage affects natural-language tasks that reuse the same computation. The failure-mode taxonomy and the component-type asymmetry should therefore be read as hypotheses to be tested on other circuits (e.g.\ factual recall, syntactic parsing), not as general principles.
 
\paragraph{Pruning protocol.}
We prune whole attention heads and MLP neurons by zeroing their weights, with no weight updates afterwards, and score importance with simple criteria (the L2 norm of a head's Q-weight slice, or of a neuron's $W_{\text{in}}$ column). We do not study weight-level unstructured pruning, removal of entire layers or blocks, gradient- or attribution-based criteria, or post-pruning fine-tuning, any of which may change or repair the failure modes we observe.
 
\paragraph{Safety implications.}
GPT-2 small has no safety fine-tuning, so our remarks on refusal- or harm-avoidance mechanisms are an extrapolation from the S-Inhibition and Negative Name Mover circuits, not a measured result.
 
\paragraph{Feature-level analysis.}
Our SAE analysis (Appendix~\ref{app:sae}) uses a SAE trained on the base model, so high reconstruction error shows only that activations leave the base model's feature manifold, not what features the pruned model has. Layer 11 is excluded because the base model's own reconstruction error exceeds the validity threshold.

\bibliography{custom}

\appendix
\section{Appendix}
\label{sec:appendix}
\subsection{Logit-Lens Divergence}
\label{app:logitlens}
We track layer-wise next-token accuracy through the logit lens on 500 TinyStories prompts, normalised by the base model's accuracy at each layer. A model is said to diverge at the first layer where the accuracy drop is significant (Bonferroni-corrected paired $t$-test, $p<0.05$) and exceeds 5 percentage points. Eight of the twelve pruned models diverge at L6 and none diverges earlier. Magnitude attention pruning diverges latest: at L9 for 30\% and 50\% sparsity (relative accuracy above 0.89) and at L7 for 70\% (relative accuracy 0.883). All three magnitude neuron models diverge at L6, with mean relative accuracy 0.670 at 30\%, whereas random neuron pruning at 30\% reaches 0.881, the same criterion reversal seen in CKA (Table~\ref{tab:fidelity}).

\subsection{Base Model DLA Reference Profile}
\label{app:base-dla}
 
Table~\ref{tab:base-dla} establishes the base GPT2 DLA profile used to compute all retention percentages reported in Table~\ref{tab:dla} of the main text. NM heads dominate the positive contribution; NNM heads make a large correct-sign negative contribution, acting as a self-correcting brake. Early-circuit classes (PT, DT, IND) have near-zero direct contributions ($|\text{DLA}| < 0.09$), consistent with their role as upstream signal processors rather than direct output contributors.
\begin{table*}[ht]
\centering
\caption{Base GPT2 DLA scores by IOI circuit class and top individual heads. Positive scores contribute towards IO over S; negative towards S (correct for NNM).}
\label{tab:base-dla}
\begin{tabular}{llrl}
\toprule
\textbf{Circuit Class} & \textbf{Class DLA} & \textbf{Top Head} & \textbf{Head DLA} \\
\midrule
Name Mover (NM)           & $+3.576$ & L9H9  & $+2.247$ \\
                          &          & L9H6  & $+0.913$ \\
                          &          & L10H0 & $+0.557$ \\
                          &          & L10H6 & $+0.208$ \\
\addlinespace
Negative Name Mover (NNM) & $-2.094$ & L10H7 & (dominant NNM) \\
\addlinespace
Backup Name Mover (BNM)   & $+0.427$ & L10H10 & $+0.455$ \\
\addlinespace
S-Inhibition (S-INH)      & $+0.384$ & L8H10  & $+0.220$ \\
\addlinespace
Previous Token (PT)       & $-0.045$ & $-$     & $-$ \\
Duplicate Token (DT)      & $-0.037$ & $-$ & $-$ \\
Induction (IND)           & $-0.085$ & $-$  &  $-$ \\
\bottomrule
\end{tabular}
\end{table*}
The retention percentage in Table~\ref{tab:dla} is defined as:
\[
  \text{Retention} (\%) = \frac{\text{DLA}_{\text{class}}^{\text{compressed}}}
                                {\text{DLA}_{\text{class}}^{\text{base}}} \times 100
\]
Negative retention values indicate the class has flipped sign relative to base and is actively harming correct IO prediction. Because NNM has a negative base DLA ($-2.094$), a \emph{positive} compressed NNM DLA produces a negative retention value, which is the inversion signature reported in Section~4.3 of the main text.
 
\subsection{Residual Path Patching: Full Four-Pathway Results}
\label{app:path-patching}
 
Table~\ref{tab:path-full} extends Table~\ref{tab:patching} by reporting path-level degradation (\%) and Cohen's $d$ for all four critical IOI communication pathways.
This provides the complete evidence base for the S-INH$\to$NM structural weak-point claim and for the circuit reorganisation signature under random neuron pruning at 30\% (negative degradation on PT$\to$IND and IND$\to$NM).
 
\begin{table*}[t] 
\centering
\caption{Residual path patching: mean degradation (\%) and paired Cohen's $d$ for four critical pathways ($n = 2{,}000$ per path per model).
Bold entries indicate the largest effect per model. Entries marked $\checkmark$ are significant ($p < 0.05$, Bonferroni-corrected). Negative degradation indicates circuit reorganisation (canonical path carries proportionally more causal weight in isolation in the compressed model).}
\label{tab:path-full}
\small
\setlength{\tabcolsep}{4pt} 
\begin{tabular}{llrrrrrrrr}
\toprule
 & & \multicolumn{2}{c}{\textbf{PT$\to$IND}} & \multicolumn{2}{c}{\textbf{DT$\to$S-INH}}
   & \multicolumn{2}{c}{\textbf{IND$\to$NM}} & \multicolumn{2}{c}{\textbf{S-INH$\to$NM}} \\
\cmidrule(lr){3-4}\cmidrule(lr){5-6}\cmidrule(lr){7-8}\cmidrule(lr){9-10}
\textbf{Spar.} & \textbf{Model} & Deg. & $d$ & Deg. & $d$ & Deg. & $d$
  & Deg. & $d$ \\
\midrule
\multirow{4}{*}{30\%}
  & Mag.\ Neu   & $255.0\%$ & $.108^\checkmark$ & $158.0\%$ & $.130^\checkmark$
                & $201.0\%$ & $.086^\checkmark$ & $\mathbf{117.1\%}$ & $.284^\checkmark$ \\
  & Rand.\ Neu  & $-96.4\%$ & $-.053^\checkmark$ & $-41.4\%$ & $-.043$
                & $-182.4\%$ & $-.100^\checkmark$ & $35.4\%$ & $.102^\checkmark$ \\
  & Mag.\ Attn  & $66.5\%$  & $.049^\checkmark$ & $62.6\%$  & $.088^\checkmark$
                & $36.6\%$  & $.027$             & $\mathbf{87.4\%}$ & $.365^\checkmark$ \\
  & Rand.\ Attn & $201.3\%$ & $.117^\checkmark$ & $122.0\%$ & $.137^\checkmark$
                & $143.2\%$ & $.084^\checkmark$ & $\mathbf{91.2\%}$ & $.310^\checkmark$ \\
\midrule
\multirow{4}{*}{50\%}
  & Mag.\ Neu   & $139.8\%$ & $.048^\checkmark$ & $112.1\%$ & $.075^\checkmark$
                & $111.2\%$ & $.039$             & $\mathbf{102.4\%}$ & $.200^\checkmark$ \\
  & Mag.\ Attn  & $128.3\%$ & $.074^\checkmark$ & $113.3\%$ & $.127^\checkmark$
                & $124.5\%$ & $.072^\checkmark$ & $\mathbf{106.8\%}$ & $.353^\checkmark$ \\
  & Rand.\ Neu  & $55.2\%$  & $.031$             & $63.5\%$  & $.068^\checkmark$
                & $20.8\%$  & $.012$             & $\mathbf{82.7\%}$  & $.261^\checkmark$ \\
  & Rand.\ Attn & $128.8\%$ & $.075^\checkmark$ & $114.7\%$ & $.130^\checkmark$
                & $128.4\%$ & $.075^\checkmark$ & $\mathbf{102.8\%}$ & $.352^\checkmark$ \\
\midrule
\multirow{4}{*}{70\%}
  & Mag.\ Neu   & $124.8\%$ & $.037$             & $107.8\%$ & $.062^\checkmark$
                & $109.9\%$ & $.033$             & $\mathbf{101.3\%}$ & $.171^\checkmark$ \\
  & Mag.\ Attn  & $144.9\%$ & $.073^\checkmark$ & $122.5\%$ & $.119^\checkmark$
                & $144.3\%$ & $.073^\checkmark$ & $\mathbf{107.7\%}$ & $.310^\checkmark$ \\
  & Rand.\ Neu  & $26.8\%$  & $.012$             & $60.6\%$  & $.053^\checkmark$
                & $20.8\%$  & $.009$             & $\mathbf{84.6\%}$  & $.217^\checkmark$ \\
  & Rand.\ Attn & $60.5\%$  & $.038$             & $79.8\%$  & $.096^\checkmark$ 
                &           &                    &           &                    \\
\bottomrule
\end{tabular}
\end{table*}
The S-INH$\to$NM pathway is the only pathway that degrades significantly under \emph{every} pruning condition, including random neuron pruning at 30\% which shows reorganisation (negative degradation) on all other pathways. This confirms the vulnerability is structural rather than a consequence of any particular importance criterion.

\subsection{Attention Pattern Analysis}
\label{app:attn-patterns}
 
\begin{table*}[t]
\centering
\caption{Attention pattern statistics under pruning, averaged over heads. $\bar H$: mean entropy (nats); Supp.: effective support $\exp(\bar H)$; $f_0$: first-token (attention-sink) attention; $\bar s$: mean per-head cosine similarity to base (Std: standard deviation across heads); $\Delta H$: mean per-head entropy change (nats). $\bar s\approx 1$ indicates routing identical to base.}
\label{tab:attention}
\small
\begin{tabular}{llrrrrrr}
\toprule
\textbf{Spar.} & \textbf{Model} & $\bar H$ & \textbf{Supp.} & $f_0$ & $\bar s$ & \textbf{Std}$(s)$ & $\Delta H$ \\
\midrule
 & Base GPT-2 & $2.013$ & $11.38$ & $.421$ & $1.000$ & -- & $0.000$ \\
\midrule
\multirow{4}{*}{30\%}
  & Mag.\ Neu   & $2.186$ & $13.47$ & $.386$ & $.775$ & $.090$ & $+0.173$ \\
  & Rand.\ Neu  & $1.638$ & $9.25$  & $.521$ & $.894$ & $.095$ & $-0.376$ \\
  & Mag.\ Attn  & $2.707$ & $28.43$ & $.270$ & $.795$ & $.330$ & $+0.694$ \\
  & Rand.\ Attn & $2.604$ & $26.27$ & $.284$ & $.704$ & $.252$ & $+0.590$ \\
\midrule
\multirow{4}{*}{50\%}
  & Mag.\ Neu   & $2.340$ & $15.92$ & $.360$ & $.706$ & $.105$ & $+0.326$ \\
  & Rand.\ Neu  & $2.050$ & $11.56$ & $.438$ & $.878$ & $.150$ & $+0.037$ \\
  & Mag.\ Attn  & $3.141$ & $40.27$ & $.178$ & $.657$ & $.310$ & $+1.127$ \\
  & Rand.\ Attn & $3.387$ & $42.14$ & $.170$ & $.587$ & $.263$ & $+1.374$ \\
\midrule
\multirow{4}{*}{70\%}
  & Mag.\ Neu   & $2.407$ & $16.79$ & $.309$ & $.625$ & $.118$ & $+0.394$ \\
  & Rand.\ Neu  & $1.478$ & $7.80$  & $.552$ & $.819$ & $.149$ & $-0.535$ \\
  & Mag.\ Attn  & $3.561$ & $51.09$ & $.098$ & $.524$ & $.253$ & $+1.548$ \\
  & Rand.\ Attn & $3.536$ & $50.35$ & $.146$ & $.482$ & $.241$ & $+1.523$ \\
\bottomrule
\end{tabular}%
\end{table*}
 
Table~\ref{tab:attention} gives the full attention statistics. Head pruning diffuses attention under both criteria (entropy rises, effective support grows, the attention sink collapses) and, for the random criterion, lowers $\bar s$ more than neuron pruning at matched sparsity ($0.704$ vs.\ $0.894$ at 30\%; $0.482$ vs.\ $0.819$ at 70\%), with larger variance across heads. Neuron pruning leaves routing comparatively intact, which is consistent with the inversion mode: circuit heads still route sensibly, but the residual stream they read from is corrupted.
 

\subsection{Activation Patching Results}
Table~\ref{tab:patching} reports head-level activation patching by functional class. S-Inhibition is the most degraded class (largest positive $d$) in all twelve conditions ($d=0.141$--$0.414$). NNM heads show a negative $d$ in every condition ($-0.205$ to $-0.539$), i.e.\ their normal suppressive contribution is decoupled or inverted rather than merely reduced.
\begin{table*}[ht]
\centering
\caption{Head-level activation patching: paired Cohen's $d$ by functional class. Positive $d$ indicates degraded causal contribution; negative $d$ indicates paradoxically larger patching effect in the compressed model (disinhibition). All overall effects significant ($p < 0.001$, Bonferroni-corrected) unless marked $\dagger$ ($p < 0.05$, uncorrected).}
\label{tab:patching}
\small
\begin{tabular}{llrrrrrrrrr}
\toprule
\textbf{Spar.} & \textbf{Model} & \textbf{PT} & \textbf{DT} & \textbf{IND}
  & \textbf{S-INH} & \textbf{NM} & \textbf{NNM} & \textbf{BNM} & \textbf{All $d$} \\
\midrule
\multirow{4}{*}{30\%}
  & Mag.\ Neu  & $.121$ & $.140$ & $.149$ & $.314$ & $.200$ & $-.341$ & $.140$ & $.135$ \\
  & Rand.\ Neu & $-.063$ & $-.037$ & $-.045$ & $.141$ & $-.016$ & $-.539$ & $-.059$ & $-.053$ \\
  & Mag.\ Attn & $.046$ & $.099$ & $.102$ & $.331$ & $.035$ & $-.467$ & $.041$ & $.053$ \\
  & Rand.\ Attn & $.107$ & $.143$ & $.153$ & $.324$ & $.160$ & $-.417$ & $.123$ & $.118$ \\
\midrule
\multirow{4}{*}{50\%}
  & Mag.\ Neu  & $.058$ & $.084$ & $.092$ & $.230$ & $.092$ & $-.244$ & $.057$ & $.074$ \\
  & Mag.\ Attn & $.078$ & $.134$ & $.161$ & $.414$ & $.113$ & $-.345$ & $.083$ & $.118$ \\
  & Rand.\ Neu & $.033$ & $.078$ & $.094$ & $.314$ & $.071$ & $-.387$ & $.028$ & $.060$ \\
  & Rand.\ Attn & $.078$ & $.137$ & $.169$ & $.402$ & $.140$ & $-.372$ & $.107$ & $.124$ \\
\midrule
\multirow{4}{*}{70\%}
  & Mag.\ Neu  & $.041$ & $.067$ & $.079$ & $.199$ & $.062$ & $-.205$ & $.041$ & $.059$ \\
  & Mag.\ Attn & $.076$ & $.127$ & $.152$ & $.366$ & $.108$ & $-.300$ & $.082$ & $.112$ \\
  & Rand.\ Neu & $.014$ & $.061$ & $.080$ & $.270$ & $.045$ & $-.329$ & $.018$ & $.047$ \\
  & Rand.\ Attn & $.042$ & $.107$ & $.136$ & $.386$ & $.097$ & $-.404$ & $.055$ & $.089$ \\
\bottomrule
\end{tabular}%
\end{table*}

\subsection{SAE Reconstruction Errors: Corroborating Evidence for Residual Stream Collapse}
\label{app:sae}
 
Sparse Autoencoder (SAE) reconstruction errors provide an independent, feature-level corroboration of the inversion mechanism described in Section~4.3. A frozen SAE trained on base GPT2 residual stream activations is applied to each compressed variant; relative reconstruction error $> 1.0$ indicates the compressed model's activations lie outside the SAE's training distribution, i.e.\ the residual stream geometry has been fundamentally disrupted.
 
\begin{table*}[ht]
\centering
\caption{SAE reconstruction errors (relative) and mean feature survival by layer across
compressed models.
$\checkmark$ = valid (relative error $< 1.0$); $\times$ = invalid (activations outside SAE
training distribution).
Layer 11 is excluded: base reconstruction error $= 2.98$ exceeds the validity threshold.
Mean survival is reported only for the highest-sparsity valid case per strategy.}
\label{tab:sae}
\small
\begin{tabular}{llrcrcc}
\toprule
\textbf{Method} & \textbf{Model} & \textbf{L1 Error} & \textbf{L1 Valid}
  & \textbf{L6 Error} & \textbf{L6 Valid} & \textbf{Mean Surv.\ (L6)} \\
\midrule
Base & GPT-2 Small (FP32) & $0.44$ & $\checkmark$ & $0.80$ & $\checkmark$ & $0.018$ \\
\midrule
\multirow{3}{*}{Mag.\ Attn}
  & 30\% & $0.44$ & $\checkmark$ & $2.44$ & $\times$ &, \\
  & 50\% & $0.44$ & $\checkmark$ & $1.28$ & $\times$ &, \\
  & 70\% & $0.35$ & $\checkmark$ & $0.38$ & $\checkmark$ & $0.0136$ \\
\midrule
\multirow{3}{*}{Mag.\ Neu}
  & 30\% & $\mathbf{155.87}$ & $\times$ & $0.67$ & $\checkmark$ &, \\
  & 50\% & $117.50$ & $\times$ & $5.27$ & $\times$ &, \\
  & 70\% & $79.20$  & $\times$ & $6.01$ & $\times$ & $0.0067$ \\
\midrule
\multirow{3}{*}{Rand.\ Attn}
  & 30\% & $2.13$ & $\times$ & $0.76$ & $\checkmark$ &, \\
  & 50\% & $0.47$ & $\checkmark$ & $20.75$ & $\times$ &, \\
  & 70\% & $0.78$ & $\checkmark$ & $0.70$  & $\checkmark$ & $0.0128$ \\
\midrule
\multirow{3}{*}{Rand.\ Neu}
  & 30\% & $0.69$ & $\checkmark$ & $16.70$ & $\times$ &, \\
  & 50\% & $0.71$ & $\checkmark$ & $33.11$ & $\times$ &, \\
  & 70\% & $0.53$ & $\checkmark$ & $0.65$  & $\checkmark$ & $0.0017$ \\
\bottomrule
\end{tabular}
\end{table*}
 
The key finding is the dissociation between SAE error and circuit head integrity under magnitude neuron pruning. At 30\% sparsity, Layer 1 reconstruction error reaches $155.87$, over 350$\times$ the base value, while all 26 IOI circuit heads remain structurally intact (weight norms $> 25$; Section~4.3).
This confirms the context-poisoning mechanism: MLP neuron removal collapses the early residual stream into a low-rank subspace that the frozen SAE cannot reconstruct, thereby corrupting the token-identity encodings that Name Mover heads read from, without physically touching the heads themselves. Random neuron pruning shows a complementary pattern: Layer 6 errors of $16.70$ and $33.11$
at 30\%--50\% sparsity, consistent with circuit reorganisation at the feature level visible in
the negative path-degradation entries of Table~\ref{tab:path-full}.

\end{document}